\documentclass{article}

\usepackage{arxiv}

\usepackage[utf8]{inputenc} 
\usepackage[T1]{fontenc}    
\usepackage{hyperref}       
\usepackage{url}            
\usepackage{booktabs}       
\usepackage{amsfonts}       
\usepackage{nicefrac}       
\usepackage{microtype}      
\usepackage{lipsum}		
\usepackage{graphicx}
\usepackage{natbib}
\usepackage{doi}

\usepackage{palatino,epsfig,latexsym}
\usepackage[acronym]{glossaries}
\usepackage{bm}

\DeclareMathOperator*{\argmin}{arg\,min}

\makeglossaries
\newacronym{EA}{EA}{evolutionary algorithm}
\newacronym{QD}{QD}{quality diversity}
\newacronym{BO}{BO}{Bayesian optimization}
\newacronym{SPHEN}{SPHEN}{surrogate-assisted phenotypic modeling}
\newacronym{SAIL}{SAIL}{surrogate-assisted illumination}
\newacronym{SADIM}{SADIM}{surrogate-assisted dimensionality mapping}
\newacronym{GP}{GP}{Gaussian process}
\newacronym{LBM}{LBM}{lattice Boltzmann method}
\newacronym{CFD}{CFD}{computational fluid dynamics}
\newacronym{MAP-Elites}{MAP-Elites}{Multidimensional Archive of Phenotypic Elites}
\newacronym{NSLC}{NSLC}{novelty search with local competition}
\newacronym{UCB}{UCB}{upper confidence bound}
\newacronym{ML}{ML}{machine learning}
\title{Efficient Quality Diversity Optimization of 3D Buildings through 2D Pre-optimization}

\date{March 27, 2023}	

\author{ 
\href{https://orcid.org/0000-0002-8668-1796}{\includegraphics[scale=0.06]{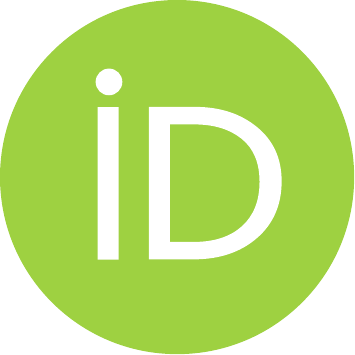}\hspace{1mm}Alexander~Hagg}\thanks{This is the final version and has been accepted for publication in Evolutionary Computation (MIT Press)} \\
\href{https://orcid.org/0000-0002-2085-7787}{\includegraphics[scale=0.06]{orcid.pdf}\hspace{1mm}\textbf{Martin~L.~Kliemank}} \\
\href{https://orcid.org/0000-0003-1133-9424}{\includegraphics[scale=0.06]{orcid.pdf}\hspace{1mm}\textbf{Alexander~Asteroth}} \\
Institute of Technology, Resource and Energy-efficient Engineering (TREE) \\
Bonn-Rhein-Sieg University of Applied Sciences \\
Sankt Augustin, 53757, Germany \\
	\texttt{alex@haggdesign.de} \\
	\And
\href{https://orcid.org/0000-0003-3263-7287}{\includegraphics[scale=0.06]{orcid.pdf}\hspace{1mm}Dominik~Wilde} \\
\href{https://orcid.org/0000-0002-9281-2027}{\includegraphics[scale=0.06]{orcid.pdf}\hspace{1mm}\textbf{Mario~C.~Bedrunka}} \\
Institute of Technology, Resource and Energy-efficient Engineering (TREE) \\
Bonn-Rhein-Sieg University of Applied Sciences \\
Sankt Augustin, 53757, Germany \\
Dpt. of Mechanical Engineering \\
University of Siegen \\
Siegen, 57076, Germany \\
    \And
\href{https://orcid.org/0000-0003-2056-6960}{\includegraphics[scale=0.06]{orcid.pdf}\hspace{1mm}Holger~Foysi} \\
Dpt. of Mechanical Engineering \\
University of Siegen \\
Siegen, 57076, Germany \\
    \And
\href{https://orcid.org/0000-0003-1480-6745}{\includegraphics[scale=0.06]{orcid.pdf}\hspace{1mm}Dirk~Reith} \\
Institute of Technology, Resource and Energy-efficient Engineering (TREE) \\
Bonn-Rhein-Sieg University of Applied Sciences \\
Fraunhofer Institute for Algorithms and Scientific Computing (SCAI) \\
Sankt Augustin, 53754, Germany \\
}

\renewcommand{\shorttitle}{Quality Diversity Pre-optimization}

\hypersetup{
pdftitle={Efficient Quality Diversity Optimization of 3D Buildings through 2D Pre-optimization},
pdfsubject={cs.NE},
pdfauthor={Alexander Hagg, Martin L.~Kliemank, Alexander Asteroth, Dominik Wilde, Mario C.~Bedrunka, Holger Foysi, Dirk Reith},
pdfkeywords={multi-solution optimization, quality diversity, efficiency, pre-optimization, wind nuisance, Bayesian optimization, lattice Boltzmann method, design process},
}

\begin{document}
\maketitle

\begin{abstract}
Quality diversity algorithms can be used to efficiently create a diverse set of solutions to inform engineers' intuition.
But quality diversity is not efficient in very expensive problems, needing 100.000s of evaluations.
Even with the assistance of surrogate models, quality diversity needs 100s or even 1000s of evaluations, which can make it use infeasible.
In this study we try to tackle this problem by using a pre-optimization strategy on a lower-dimensional optimization problem and then map the solutions to a higher-dimensional case.
For a use case to design buildings that minimize wind nuisance, we show that we can predict flow features around 3D buildings from 2D flow features around building footprints.
For a diverse set of building designs, by sampling the space of 2D footprints with a quality diversity algorithm, a predictive model can be trained that is more accurate than when trained on a set of footprints that were selected with a space-filling algorithm like the Sobol sequence.
Simulating only 16 buildings in 3D, a set of 1024 building designs with low predicted wind nuisance is created.
We show that we can produce better machine learning models by producing training data with quality diversity instead of using common sampling techniques.
The method can bootstrap generative design in a computationally expensive 3D domain and allow engineers to sweep the design space, understanding wind nuisance in early design phases.
\end{abstract}


\section{Introduction}

Early design decisions in the built environment have a large impact on the feasibility of designs.
Building regulations often prevent building designs from being accepted, for example, due to limits on the maximum allowed wind nuisance around the design.
Wind nuisance is determined by the maximum flow velocity ($\mathbf{u}_{max}$) in the vicinity of the building \citep{NEN8100}.
If the maximum allowed nuisance is breached, the design has to be adapted, which necessitates expensive adaptation phases, often times performed in a wind tunnel, or can cause up to major revisions of a design.
Some of the overarching goals in the generative design for the built environment include efficiently finding a wide variety of good building designs from scratch to enhance engineers' knowledge early on in the architectural design process, preventing design mistakes that have a negative impact on the quality of solutions.
In the past decade, novel multi-solution optimization methods like \gls{QD} have been introduced.
These methods create large archives of solutions by using phenotypic/behavioral feature spaces for niching, see \citep{Hagg2020c}.
However, applying these methods to expensive 3D fluid dynamics domains is not always feasible.
The computational cost of calculating $\mathbf{u}_{max}$ for all possible design variations early on is prohibitively large, as it might take hundreds of thousands of evaluation.
Even though the use of surrogate models reduces the number of required evaluations to hundreds or thousands, shown by \citep{gaier2018data} and \citep{Hagg2020a}, this number is too large when simulations can take hours or even days.
Yet improving understanding of fluid dynamics in the design of the built environment during the initial design phase is paramount to the success and efficiency of the entire design process.

\begin{figure}
    \centering
    \includegraphics{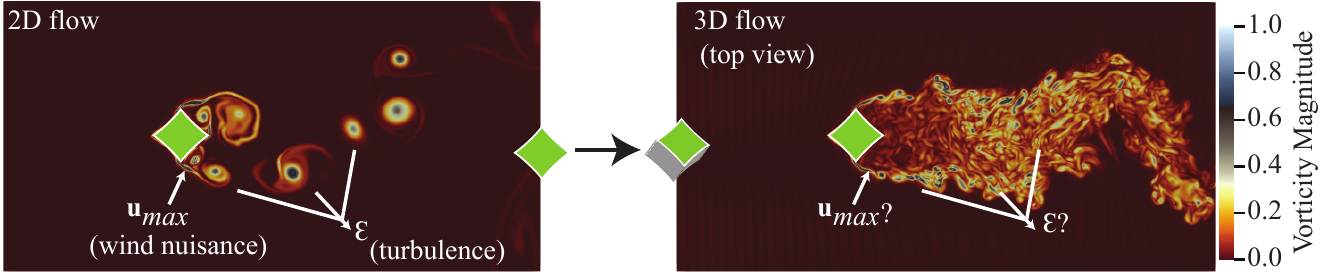}
    \caption{3D flow is qualitatively different from 2D flow, shown by the difference in the flow of the 2D building footprint (left) and an extruded version of that shape in 3D (right). Can we map flow features like wind nuisance and turbulence metrics of 2D footprints to those of 3D building designs? }
    \label{fig:2D3Ddifference}
\end{figure}

The flow around a building is usually unsteady and optimization of steady flows may result in misleading interpretations, as observed in \citep{blocken2016pedestrian}. 
3D flow is qualitatively different from 2D flow in terms of turbulence development, shown in \citep{Boffetta2012, ecke2017}. 
Figure~\ref{fig:2D3Ddifference} depicts the vorticity field of a flow simulation with an exemplary shape in 2D and 3D. 
Turbulent flow features differ significantly in two and three dimensions, due to the fact that 2D flows exhibit an inverse energy cascade to larger scales and a direct enstrophy cascade to smaller scales. 
Although a direct energy cascade to smaller scales is usually reported in 3D flows, an inverse cascade can be observed, too, when mirror symmetry is broken, as observed in \citep{Biferale2012}.
These differences can lead to different conclusions depending on dimensionality. \citet{trizila2008surrogate} showed, that although the trend of some flow features (lift and plunging amplitude) can be similar, the trend of the phase lag changes significantly. 
\citet{font2021deep} showed that the flow around 3D cylindrical objects can be predicted based on spanwise-averaged Navier-Stokes equations. 
However, it is unclear how their method handles the wide diversity of shapes in early design stages.
To perform the \gls{CFD} simulations, the \gls{LBM} has made considerable progress in the prediction of turbulent flows, including direct numerical simulation (DNS), as in \citep{falcucci2021}, under-resolved DNS, as in \citep{dorschner2016}, and large eddy simulations (LES), as in \citep{buffa2021}.
The \gls{LBM} is renowned for its high performance, shown in \citep{Lallemand2021}, although 3D simulations remain a costly endeavor. 
This increases the demand for efficiency of \gls{QD} methods applied in these kinds of domains even more.

Generative design and \gls{ML} methods are increasingly being established in fluid dynamics, for example in \citep{lye2020deep}, helping engineers to better understand their problem domains earlier on in the design process, and increase the likelihood of making the right design decisions. 
These methods generally need large data sets with a wide variety of shapes and flow patterns.
Although novel methods like physics-informed neural networks require fewer data points in some cases, for example in work by \citet{sun2020surrogate}, it is unclear how they perform on the wide variety of shapes in early design stages.
The increased integration of \gls{ML} in the design process is pushing methods in fluid dynamics and optimization to their limits, due to the requirements on the amount and variation of training data points.
Especially when data sets are unavailable for a certain problem, they need to be created first.
We examine this problem of bootstrapping the data creation process with \gls{QD}.

As 2D flow simulations are several orders of magnitude cheaper than 3D simulations, the prior optimization of 2D replica can be a cost-effective strategy.
Pre-optimization in 2D is an efficient way to gain an initial understanding of high-performing shapes and reduce the solution space, as was done by \citet{zhong20173d}, reducing the number of expensive 3D simulations necessary for optimization.
This work sets out to use 2D pre-optimization for 3D optimization in the context of the solution diversity created by \gls{QD} algorithms.
As \gls{CFD} simulation of 3D domains is much more computationally expensive than 2D, a pre-optimization step greatly increases the applicability of \gls{QD} in 3D fluid dynamics domains.

To map 2D pre-optimized shape sets to 3D, this work analyses how accurately we can predict flow features around a diverse set of extruded 3D shapes from their 2D counterparts. It is obvious that often 3D shapes are not just extruded versions of a 2D shape. Still, whenever 3D shapes can easily be derived from a 2D footprint, such as in the case of flows around axial objects like high-rise buildings, a prior 2D optimization potentially decreases computational costs in 3D shape optimization.
Based on these 3D extrusions, promising shapes can be identified upon which further optimization in 3D may be required, although the latter step is not part of the scope of this work.

\gls{QD} was shown to be a good bootstrapped data generator for \gls{ML}, for example by \citet{gaier2018data}, \citet{gaier2020discovering}, \citet{hagg2021expressivity}, \citet{bentley2022coil}, and \citet{bossek2022exploring}.
We, therefore, present a method that maps features of 2D to 3D designs with surrogate-assisted \gls{QD}, which serves both as an optimization strategy as well as a data generator for \gls{ML}.
We hypothesize that \gls{QD} can be used as a pre-optimizer in mapping optimal 2D to 3D building designs.
If we can confirm this hypothesis, we can build models that can predict 3D flow features based on 2D flow features alone, for extruded 3D shapes.
This benefit would enable us to run a computationally less expensive optimization using 2D flow simulations to produce an initial building design set for 3D flow optimization.
Being able to predict certain key 3D flow features efficiently provides architects and engineers with the ability to make more informed early design decisions. 
The following research questions are formulated:
\begin{enumerate}
    \item Can we produce better ML models by producing training data with \gls{QD} instead of common sampling techniques?
    \item Can we map high-performing 2D to 3D shapes using \gls{QD}?
\end{enumerate}

This work is structured as follows. Section \ref{sec:sadim} describes the methods that are used to find diverse, high-performing shapes and generate training data that allows the prediction of 3D from 2D flow features. 
In section \ref{sec:evaluation} we evaluate whether we can accurately predict 3D flow features for extruded shapes and how this can be used to find a large set of high-performing building designs.
Finally, we draw conclusions in section \ref{sec:conclusions}.

\section{Methodology}
\label{sec:sadim}

In order to map solutions in a 2D domain to a 3D domain, we ask ourselves whether we can predict the maximum flow velocity ($\mathbf{u}_{max}$) and average enstrophy ($\mathcal{E}$), a measure for turbulence, in 3D flow fields, based on their 2D counterparts.
We utilize a \gls{QD} algorithm to find diverse, high-performance building footprints in a 2D air flow (see section \ref{sec:optimization:qd}). 
The 2D building footprints are extruded into 3D building designs, after which 3D flow features are predicted from the former 2D footprints.
We hypothesize that the optimization method creates a sample set that we can model more easily and accurately than a sample set created using a Sobol sequence.
Sobol sequences \citep{sobol1967distribution} are space-filling pseudo-random sequences that guarantee that the samples are spread out evenly, something that a random distribution cannot guarantee.

\begin{figure}[ht]
    \centering
    \includegraphics{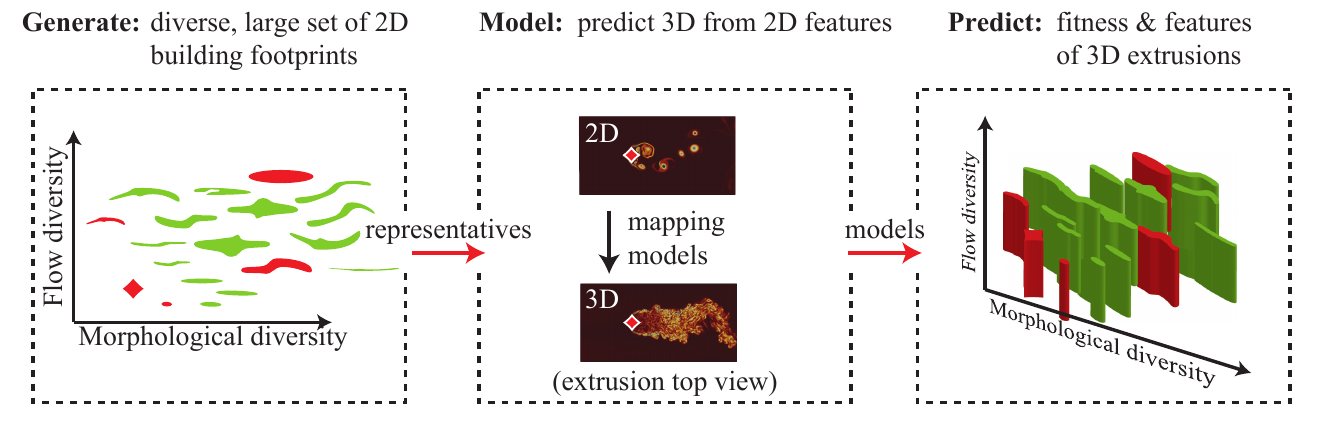}
    \caption{\textbf{\Glsdesc{SADIM} (\gls{SADIM})}. By efficiently creating a large and diverse, high-performing set of 2D building footprints (\textbf{Generate}), then simulating only a representative selection of shapes in 3D and creating a predictive model (\textbf{Model}), a prediction can be given of the flow features around 3D extrusions (\textbf{Predict}). In \textbf{Generate}, the diversity of the shape set is forced over the feature dimensions of the diversity of flow (enstrophy $\mathcal{E}$) and morphological diversity ($\mathbf{area}$). Solution fitness is determined by the maximum flow velocity $\mathbf{u}_{max}$ caused by the shape.}
    \label{fig:mapping}
\end{figure}

Predicting features of 3D flow from 2D for a random set of shapes is not expected to be accurate, due to the physical differences in both flow regimes.
However, if we were to be able to evolve a set of shapes for which we can predict 3D flow features, we could use the 2D flow regime as a pre-optimization step for 3D optimization.
The main idea is to create a set of 2D shapes that is diverse enough to capture sufficient information about the differences between the flow around those 2D shapes and 3D extrusions of those shapes.
The method consists of three steps, shown in figure~\ref{fig:mapping}. 

An efficient optimization algorithm is used to create a diverse set of 2D building footprints (\textbf{Generate}). 
From this set, representative shapes are selected and a \gls{CFD} solver is used to simulate the 2D flow (\textbf{Model}). 
The selected shapes are extruded and 3D flow is calculated, after which relevant flow features are extracted.
A predictive (surrogate) model is trained that can map 2D to 3D flow features.
Finding such a surrogate model would allow us to pre-optimize 3D shapes in 2D first, at least for the case where we extrude the 2D shapes into 3D.
In the last step, the flow features of the 3D extrusions of shapes discovered in I are predicted to be able to build the shape archive of 3D extrusions (\textbf{Predict}).
Thereafter, the extrusions of the optimized 2D shapes serve as an initial set for full 3D optimization.

We aptly named this methodology \gls{SADIM}, which is akin to space mappings in which coarse models are mapped to expensive, fine models. 
\Gls{SADIM} specifically targets coarse models that are of lower dimensionality than the final target model, in this case, 3D fluid dynamics. 
\gls{SADIM} allows us to increase our understanding of how 2D flow around high-performing shapes transfers to the 3D domain. 
This can help us to efficiently create diverse sets of solutions in very expensive 3D domains.
\gls{SADIM} is a step towards efficient \gls{QD} for 3D fluid dynamics problems, although the complete 3D optimization is not within the scope of this work. 

The following parts discuss the \gls{QD} optimization method used in (\textbf{Generate} and \textbf{Predict}), the fluid dynamics method that is used in (\textbf{Generate} and \textbf{Model}), and the selection method used in (\textbf{Generate}).

\subsection{Quality Diversity Optimization}
\label{sec:optimization:qd}
Shape optimization problems are characterized by many degrees of freedom with only a few flow variables to be optimized. 
Common variables comprise drag, maximum velocity, or turbulent properties. 
A performance function $f(\mathbf{x})$ that formalizes the optimization goal is defined based on those variables.
The (unconstrained) optimization problem is therefore defined as follows:
\begin{equation} \label{eq:optimization}
\mathbf{x}_{min} = \argmin_\mathbf{x} (f(\mathbf{x}))
\end{equation}
where $\mathbf{x}_{min}$ is the global minimizer of the function $f(\mathbf{x})$. In case the function needs to be maximized, the equation has to be adapted accordingly.

If possible, the gradient of the flow variables with respect to the shape parameters is determined to perform the optimization. 
The optimization iteratively leads the path toward an optimized shape, in the direction of this gradient, if available.
While customary gradient-descent methods are usually avoided due to numerical costs, as established in \citep{Jahangirian2011}, the most widely used approach for the gradient calculation are adjoint-based methods, for example in \citep{Pironneau1974,Marinc2012}. 
Adjoint-based methods were established in \gls{CFD} by \citet{Jameson1988}.
They demonstrated the optimization of a 3D aircraft wing using the Navier-Stokes equations. Later these methods were also applied using \gls{LBM} by \citet{Krause2013}. 
All of these gradient-based approaches have the disadvantage that the found solution may only be a local optimum, requiring further measures, and that design considerations are solely a secondary matter. 

By contrast, \glspl{EA} are a viable alternative that attracted attention also for \gls{CFD} applications in the last decades, examples of which can be found in \citep{Foli2006,Jahangirian2011}. 
\Glspl{EA} do not use an explicit gradient of the performance function to infer the direction in which the optimum is located. 
They use population-based sampling and approximate the gradient by ranking the samples. 
The search is then driven by selecting samples based on their relative performance ranking and adding perturbations to keep the population moving in the direction of the inferred gradient.
\Glspl{EA} have long been at the forefront of optimization domains that were either too expensive or not well understood to find solutions in analytical ways. 
Not needing an explicit gradient, EA have been shown to find unexpected solutions by \citet{lehman2020surprising}. 
But with analytical methods coming out strong in physics domains like fluid dynamics, EA are not used in single-solution optimization settings as often anymore.
However, performance is not the only goal in real-world optimization. 
Putting a focus on the diversity of solutions, and finding more than one way to solve a problem, allows designers and engineers to use optimization algorithms much earlier in the design process. 
Multi-solution optimization is a field that is getting more attention, due to the advent of generative models and \gls{QD} algorithms.
The multi-solution optimization problem is defined as follows:
\begin{equation} \label{eq:multioptimization}
\mathbf{X}_{min} = \argmin_\mathbf{x} (f(\mathbf{x})), |\mathbf{x}_i - \mathbf{x}| \le \epsilon
\end{equation}
where $\mathbf{X}_{min}$ is the set of solutions that minimizes $f(\mathbf{x})$ in a local neighborhood, a \textit{niche}, defined by $\epsilon$ on the parameters.
Niching, which was inspired by biological evolution, retains solutions that might not perform as well as others, but whose distance to better solutions is large enough.
Novel solutions are thus retained in the solution set, even if they are outperformed by others.
\citet{Hagg2020b} gave evidence that multi-solution algorithms provide a more diverse solution set than classical approaches. 

\gls{QD} algorithms combine performance with \textit{novelty search}, which was introduced by \citet{Lehman2011a} and ignores performance altogether. 
\gls{QD} \citep{pugh2015confronting} finds a diverse set of solutions by using a concept called \textit{phenotypic niching}. Instead of measuring the distance between solutions in parameter space, phenotypic niching measures the similarity between solutions based on certain morphological or behavioral features that can usually only be measured in expensive simulations.
\gls{QD} optimization solves the problem
\begin{equation} \label{eq:qd}
\mathbf{X}_{min} = \argmin_\mathbf{x} (f(\mathbf{x})), |p(\mathbf{x}_i) - p(\mathbf{x})| \le \epsilon
\end{equation}
where $\mathbf{X}_{min}$ is the set of solutions that minimizes $f(\mathbf{x})$ in a local neighborhood, a \textit{niche}, defined by $\epsilon$ on the phenotypic features considered by the function $p(\mathbf{x})$.

\paragraph{Voronoi-Elites}
One of the first and most widely used \gls{QD} algorithms, \gls{MAP-Elites} \citep{mouret2015illuminating,cully2015robots}, fills a rectangular grid of fixed niches with high-performing solutions.
The main disadvantage of using an archive with fixed niches lies during the initialization of the archive.
If the solutions are phenotypically very similar, local competition can greatly reduce the genetic variance of the initial population.
Instead of using a fixed archive, in this work we use the Voronoi-Elites algorithm \citep{Hagg2020b}, which uses an archive with dynamic niches, similar to \gls{NSLC}~\citep{lehman2011evolving}\footnote{Code available at \url{https://github.com/alexander-hagg/v-elites}}.

\begin{figure}
    \centering
    \includegraphics{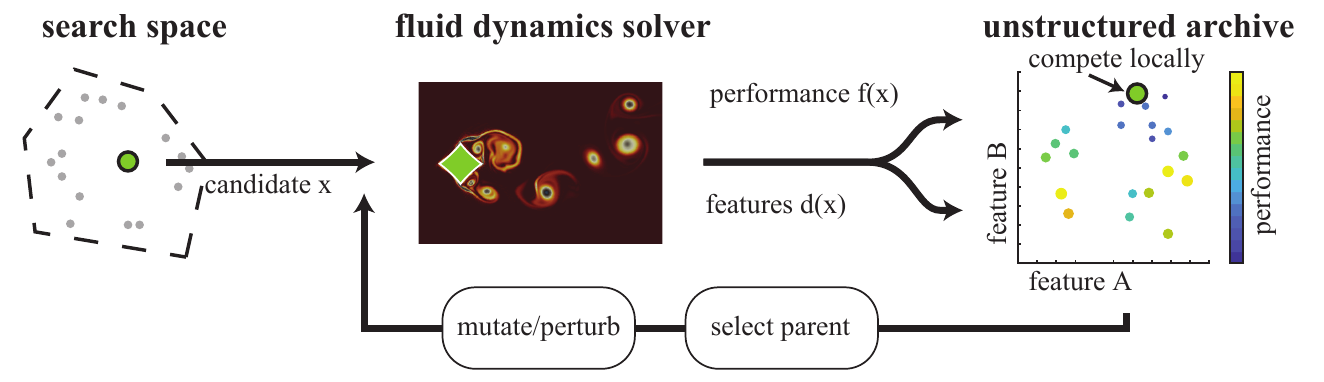}
    \caption{Voronoi-Elites searches in parameter space but performs niching in an unstructured archive defined by phenotypic descriptors, like morphological and flow features. Candidate solutions are assigned to a location in the archive based on their feature values. If the archive reaches its maximum size, the closest pair of solutions compete to stay in the archive.}
    \label{img:archive}
\end{figure}

Voronoi-Elites performs a search in parameter space (see figure~\ref{img:archive}).
Solutions are decoded into their phenotypes - full descriptions of shapes.
Their performance is determined, for example by placing the phenotypes in a fluid dynamics simulation. 
From here, secondary flow features can be calculated alongside the performance value of the phenotypes. 
In fluid dynamics, we are interested in finding diverse solutions corresponding to certain flow characteristics, like enstrophy and maximum flow velocity. 
An archive is used to keep track of performance and diversity, based on the secondary flow features. 
Solutions are added to the archive, but as soon as the archive's size increases past the maximum size, the pairs of solutions that are closest in feature space are compared.
The worst solution of the closest pair is ejected from the archive. 
Over a fixed number of iterations, new candidate solutions are generated by randomly selecting parents from the archive and perturbing their genome by drawing a perturbation from a normal distribution.
Voronoi-Elites produces diverse, large sets of high-performing solutions, yet it oftentimes needs hundreds of thousands of evaluations and is inefficient.
The algorithm is called Voronoi-Elites because the visualization of the archive defines a Voronoi diagram.

\paragraph{Surrogate Assisted Quality Diversity}

While \gls{QD} produces highly diverse solution sets, it needs too many evaluations to be used in expensive domains like fluid dynamics. 
It is infeasible to apply \gls{QD} to the fluid dynamics domain without the use of surrogate models. 
Surrogate assistance replaces most evaluations by a statistical model. 
In \gls{BO}, \gls{GP}regression models, which were introduced by \citet{rasmussen1997}, are commonly used. 
\gls{GP} models predict the performance of new solutions based on the parameter-wise distance to known examples. 
Solutions whose parameters are close together should have a higher correlation. 
This assumption is modeled using a covariance function, with the additional assumption of smoothness.
One popular example of such a function is the squared exponential covariance function $k$, which is defined as follows:

\begin{equation}
k(x,x') = \sigma^2 \cdot exp\left(-\frac{{(x - x')}^2}{2l^2}\right)
\label{eq:covariance}
\end{equation}

It has two hyperparameters: the length scale $l$, which determines the sphere of influence of known examples, and the signal variance $\sigma$, which is determined by the scale of the function. 
If these hyperparameters are not known for a domain, they are usually determined by minimizing the negative log-likelihood of the process. 
The covariance function $k$ and a set of known examples $x$ constitute the \gls{GP}, which is used to predict a distribution per location. 
Besides the mean value $\mu$ prediction for a location, the \gls{GP} also outputs a confidence interval, designated with $\sigma$ for that prediction.
$\sigma$ tends to be high for locations far away from known examples. 

Instead of using only $\mu$, the sampling or \textit{acquisition function} also takes into account the uncertainty $\sigma$ to provide a more efficient sampling method. One commonly used acquisition function is \gls{UCB}, which was formulated by \citet{auer2002}.

\begin{equation}
\text{UCB}(x) = \mu(x) + \kappa \cdot \sigma(x)
\label{eq:ucb}
\end{equation}
The parameter $\kappa$ tunes the \gls{UCB} function between exploitation ($\kappa = 0$) and exploration ($\kappa \gg 0$). 
By using a high $\kappa$, the resulting optimized samples are more prone to be located in high-performing unknown regions of the modeling space. 
UCB proved to be an excellent sampling mechanism for multi-solution optimization.

The first Bayesian \gls{QD} method using \gls{UCB} was surrogate-assisted illumination (SAIL) by \citet{gaier2018data}. 
It combined \gls{BO} with the \gls{MAP-Elites} \citep{mouret2015illuminating,cully2015robots} algorithm.
They used \gls{UCB} as an optimistic sampling method on a \gls{QD} archive to find examples that were not only located near the global optimum but over an entire range of different solutions. 
A surrogate model was used to predict solution performance and to greatly increase the efficiency of this QD method. 
This Bayesian interpretation of \gls{QD} uses its own feature-based archive to guide the sampling of the simulation. 
Extending SAIL, \citet{Hagg2020a} gave evidence that \gls{GP} models can even predict features alongside their performance. 
Their method, \gls{SPHEN}\footnote{Code available at \url{https://github.com/alexander-hagg/sphenpy/}}, efficiently produced high-performing solution sets with a large diversity of flow patterns around 2D shapes. 

\begin{figure}
    \centering
    \includegraphics{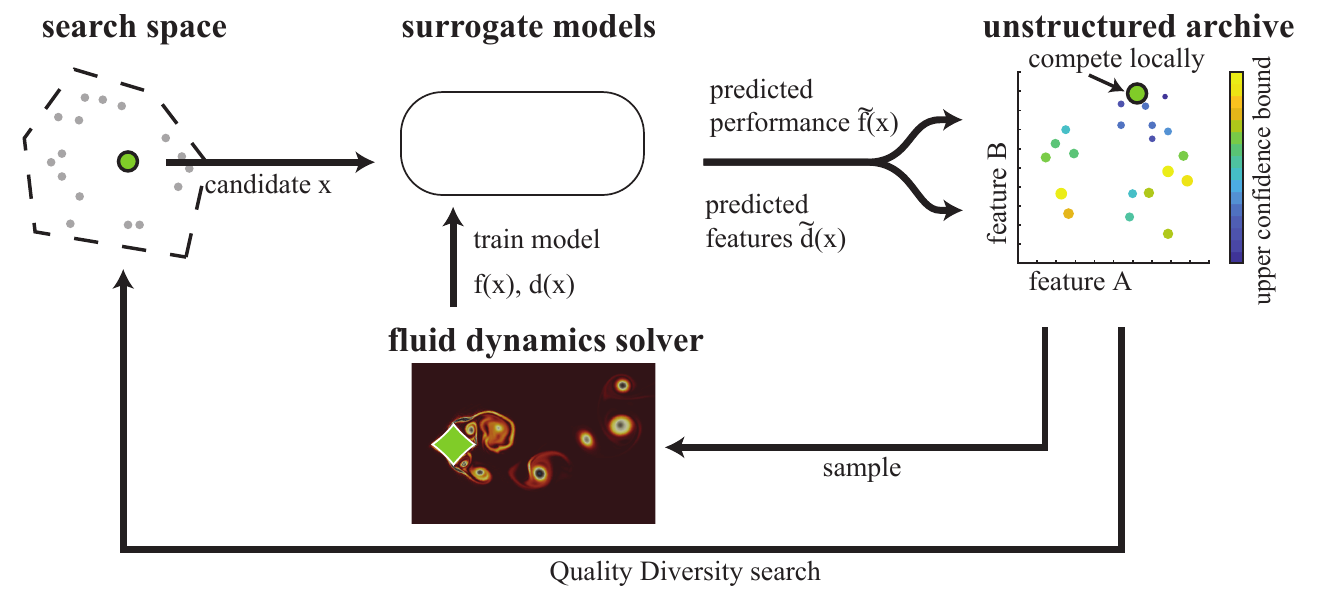}
    \caption{In \gls{SPHEN}, the performance and high-level features (flow and morphology) are determined for an initial random set of design parameter tuples. Surrogate models are trained for both performance $\widetilde{\text{f}}$(x) and feature descriptors $\widetilde{\text{d}}$(x). The models are then used to illuminate the design space and determine the best locations for further sampling. The loop is continued until a certain budget is reached. After this, the models are used to produce a final archive of solutions.}
    \label{img:sphen}
\end{figure}

Figure~\ref{img:sphen} shows how \gls{SPHEN} fills a Voronoi-Elites niching archive based not upon performance alone but on the optimistic prediction of performance, the \gls{UCB}. 
The resulting archive contains candidate samples, which are selected at random or using a space-filling method. 
A number $n$ of samples is extracted from the archive and evaluated to form an additional set of training samples. 
In SPHEN, both the performance as well as the features of a solution are modeled with \gls{GP}. 
After updating the \gls{GP} model(s), the archive is filled again until the computational budget is exhausted. 
By setting the $\kappa$ parameter in \gls{UCB} to $0$, the resulting \gls{GP} model can now be used to fill the final archive.
The archive now contains solutions only based on performance and features, but using only a small number of direct samples from simulation. 
The \gls{GP} models describe the performance and flow features of solutions within an \textit{elite hypervolume}, a term coined by \citet{Vassiliades2018}, a set of (possibly disjoint) high-performance regions in parameter space. 
SPHEN enables the use of SAIL with expensive-to-calculate flow features. 
It can produce solution sets with a large variety of flow features and good performance. 

\subsection{Selecting Representatives}
In order to use \gls{SPHEN} in real-world applications, it still needs hundreds or even thousands of simulations, which is infeasible for many 3D domains.
\gls{SADIM} therefore only selects a small set of representative shapes to create a model that is used to predict 3D features from 2D counterparts.
We reduce the maximum size of the \gls{QD} archive to select a small number of representative 2D solutions (see the \textbf{Generate} and \textbf{Model} steps in figure \ref{fig:mapping}).
A new archive is created based on the same features, but with a reduced maximum size.
When the members of the old archive are now added to the new, reduced archive, the highest-performing representatives of the diverse set are automatically selected.

\section{Evaluation}
\label{sec:evaluation}
As explained in Section~\ref{sec:sadim}, we use \gls{SPHEN}, a surrogate-assisted \gls{QD} method, to generate a diverse set of optimized 2D shapes. 
We predict their 3D counterparts' flow features using only a small number of 3D simulations.
The working hypothesis is that the flow features of 2D shapes sampled with \gls{SPHEN} can be more accurately predicted than randomly sampling 2D shapes.
The modeling problem should be easier, because \gls{QD} algorithms produce optimized shapes that are part of the elite hypervolume, shown by \citep[][]{Vassiliades2018}, due to symmetries and similarities between the shapes in that region of the search space.
This work tests whether we can indeed accurately rank 3D shapes based on their 2D counterparts despite the systematic discrepancy between 2D and 3D flow features.

Since \gls{EA} compare solutions based on their relative ranking only, a surrogate model used in this context is assumed to be of good quality if the rank of selected individuals coincides with the rank based on the real simulation. An extensive review of surrogate-assisted \gls{EA} can be found in \citep{jin2011surrogate}.
To determine the quality of the model under this aspect, the ranking accuracy of \gls{GP} models trained on a random selection of shapes is compared to models trained on a set of shapes pre-optimized by \gls{SPHEN} in 2D.

\subsection{Experimental Setup}
The optimization domain, configuration of \gls{SPHEN} and \gls{SADIM} algorithms, and some particularities of the flow simulation are explained here.

\subsubsection{Architecture Domain}
\label{sec:evaluation:domain}
Novel design techniques in the built environment allow architects to design 3D buildings even more freely.
The selection of feasible designs however is constrained by the airflow around buildings, due to restrictions on the maximum wind nuisance they may cause.
Building regulations defined in \citep{NEN8100} assume a main or common wind velocity and measure the maximum flow velocity $\mathbf{u}_{max}$ in a large area around the building.
Turbulent flows are common around buildings, evident from \citep{blocken2016pedestrian}. 
This is why it makes sense to resolve turbulence differences between solutions through the use of enstrophy $\mathcal{E}$ as a feature. 
Higher enstrophy is equivalent to stronger turbulence around a building.

\begin{figure}
    \centering
    \includegraphics{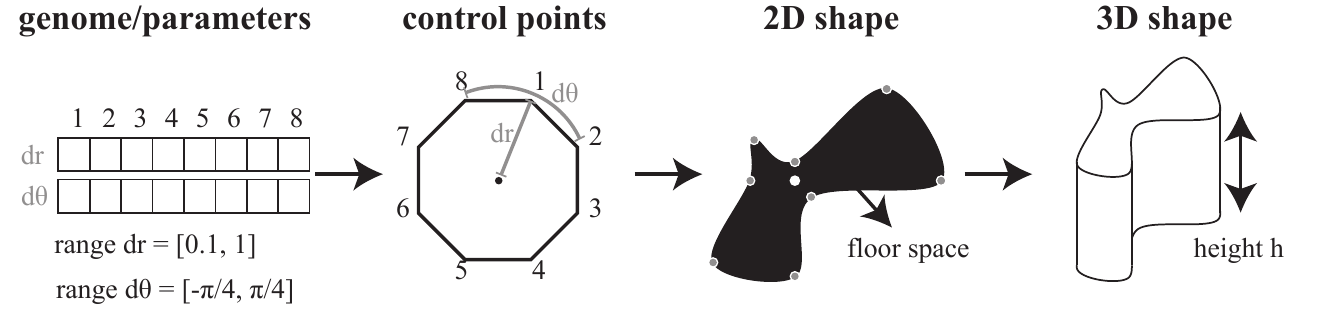}
    \caption{Shapes are encoded with 16 polar control points that get transformed into the final shape using locally interpolating splines (Catmull-Rom). The shape is then extruded to serve as the 3D shape for the numerical simulation.}
    \label{img:encoding}
\end{figure}

The problem domain is a simplification of common 3D architecture and is based on a 2D shape domain, which was used by \citet{Hagg2020b}. 
Solutions consist of local interpolating Catmull-Rom splines, a representation that was introduced by \citet{catmull1974}. 
The splines are encoded by a polar coordinate-based genome (see figure~\ref{img:encoding}). 
The polar coordinates of the polygon's control points are controlled with the radius $r$ and angle $\theta$. 
Using the control points and Catmull-Rom splines, a large variety of smooth convex and concave shapes can be created. 
The 2D shapes are mapped to 3D by a simple extrusion method and a fixed height $h$. 
The optimization process aims to minimize $\mathbf{u}_{max}$ while varying the floor space $\mathbf{A}$ and enstrophy $\mathcal{E}$.

\subsubsection{Configuration of SPHEN and SADIM}
An initial sampling set of 64 2D building footprints is created using a Sobol sequence. 
The shapes are simulated in Lettuce to extract $\mathbf{u}_{max}$ and $\mathcal{E}$.
Internal \gls{GP} surrogate models are trained to predict these flow features from the shape parameters. 
The \gls{GP} models use the (isotropic) squared exponential function described in equation \ref{eq:covariance} to estimate distances between samples. Its hyperparameters (see section \ref{sec:optimization:qd}), length scale $l$, and signal variance $\sigma$, are determined by minimizing the log-likelihood with the internal exact inference method of the GPML library, written and maintained by \citet{rasmussen2010gaussian}.

\gls{SPHEN} uses the surrogate models to efficiently discover a diverse set of high-performing shapes.
The solutions are saved into an archive, defined by two dimensions: area $\mathcal{A}$ of the shape (floor space of the building) and $\mathcal{E}$ of the flow. 
Solutions are selected based on whether they minimize $\mathbf{u}_{max}$.
The surrogate models are continuously trained within \gls{SPHEN}, selecting 10 new samples after every round of the internal \gls{QD} search.
100 iterations of SPHEN's outer loop are executed to retain 1000 new samples from the 2D Lettuce simulation.
In each iteration, a full \gls{QD} optimization with 1024 generations is executed. 
The archive is quickly filled by randomly selecting 32 samples from the archive and perturbing their parameters with a value from a Gaussian distribution with $\sigma = 0.1$. 
We fill the archive until a maximum number of solutions (1024) is reached.
After reaching a filled archive, any new candidate solution is compared to the archived solution that has the most similar (predicted) features. 
Between the two, the solution with the lowest predicted $\mathbf{u}_{max}$ is kept in the archive and the other is rejected. 
The \gls{GP} models are retrained after every iteration by selecting 10 evenly spread members of the archive using a Sobol sequence on the archive's dimensions. 
These new samples are then simulated in Lettuce and added to the training set of the \gls{GP} models. 
Starting from 64 samples, the training set grows until 1000 samples are available.

This is all done to create the high-performing diverse 2D set described in the \textbf{Generate} step of figure~\ref{fig:mapping}. 
Then we select 16 representative samples from the 1000 available ones, run a full 3D simulation of the flow around the corresponding extrusion to determine 3D flow characteristics, and train a mapping surrogate that maps 2D features to 3D features (\textbf{Model}). The surrogate is configured similarly to the internal \gls{GP} models in \gls{SPHEN} (see section \ref{sec:optimization:qd}).
The full archive that was created by \gls{SPHEN} is now mapped to a 3D archive using newly trained \gls{GP} models (\textbf{Predict}).

\subsubsection{Fluid Dynamics Solver}
\label{sec:fluidsolver}
This section recapitulates the \gls{LBM}, which is used for the fluid dynamics simulations in this study. 
The simulations are performed by the Python-based \gls{LBM} framework Lettuce, which was introduced in \citep{lettuce} and is based on PyTorch and thus allows hardware accelerated simulations on GPUs\footnote{Code available at \url{https://github.com/lettucecfd/lettuce}}. 
The \gls{LBM}, based on work in \citep{mcnamara1988, Kruger2016, Lallemand2021}, is an established method for the simulation of fluid dynamics. 
Emerging from kinetic theory, the \gls{LBM} describes the evolution of fluids on a mesoscopic scale in terms of discrete particle distribution function $f_i (\bm{x},t)$ in space and time by the lattice Boltzmann equation
\begin{equation}
    f_{i}\left(\mathbf{x}+\mathbf{c}_{i}\delta_{t}, t+\delta_t \right) = f_{i}\left(\mathbf{x},t \right) + \Omega_{i}\left(\textbf{x},t\right),
\end{equation}
with the discrete collision operator $\Omega_i$. 
The variables $\mathbf{c}_i$ and $\delta_t$ define the discrete particle velocities and time step, respectively. 
This equation yields the algorithm of the \gls{LBM}, which in its most basic form consists of two operations:
\begin{enumerate}
\item The \emph{collision} step models the effects of particles colliding elastically with each other by $\Omega_i$.
\item The \emph{streaming} step moves all particles according to $\mathbf{c}_{i} \delta_{t}$.
\end{enumerate}
The well-known D2Q9 stencil with nine discrete velocities and the D3Q27 stencil with 27 discrete velocities is used in two and three dimensions for the velocity space, respectively. 
The macroscopic variables of interest, in particular the density $\rho$ or the momentum of the fluid $\rho\textbf{u}$, can be computed directly from the particle distribution $f_i$,
\begin{equation}
    \rho \left( \mathbf{x}, t\right) = \sum_{i} f_{i} \left(\mathbf{x},t \right) 
    \qquad \mathrm{and} \qquad 
    \rho \mathbf{u}\left( \mathbf{x}, t\right) = \sum_{i} \mathbf{c}_{i}f_{i} \left(\mathbf{x},t \right).
\end{equation}
The macroscopic values used in the optimization process are the maximum of the velocity $\mathbf{u}_{max}$ and the integrated enstrophy $\mathcal{E}$ in $D$ dimensions within the flow domain $\Gamma$

\begin{equation}
    \mathcal{E} = \frac{1}{V(\Gamma)} \int_\Gamma \ (\nabla \times \mathbf{u})^2 \  d^D \mathbf{x},
\end{equation}

where $V(\Gamma)$ is either the domain area (floor space) in 2D or the domain volume in 3D. The enstrophy is used as an indicator for turbulence since it measures the vorticity in the flow domain $\Gamma$.

\paragraph{Collision Operator}
A common choice for the collision operator is the Bhatnagar-Gross-Krook (BGK) operator by \citet{Bhatnagar1954}, which is straightforward but also prone to instabilities for large Reynolds numbers, evident in \citep{Nathen2019, Kramer2019}. 
To target this issue, many alternative collision models have been introduced such as multi-relaxation time models in \citep{Lallemand.2000,Dellar.2003}, two-relaxation time models in \citep{Ginzburg2008}, regularized models in \citep{Latt2006}, entropic models in \citep{Karlin1999} or cumulant models in \citep{Geier2015}.
In this work, the entropic multi-relaxation time model by Karlin-Bösch-Chikatamarla (KBC) by \citet{Karlin2014} is used, which proved to be suitable to simulate under-resolved, turbulent flows in several works by \citet{Dorschner2017,DiIlio2018,Kocheemoolayil2019,Kramer2019}.

At its core, the KBC collision operator splits the distribution functions into
\begin{equation}
    f_i = k_i + s_i + h_i,
\end{equation}
where $k_i$, $s_i$, and $h_i$ describe the conserved kinetic moments, the shear moments, and the non-physical higher-order moments, respectively. 
The collision operator relaxes these moments towards their individual equilibria according to
\begin{equation}
    \Omega_i(f_i) = -\beta(2 (s^\mathrm{neq}_i ) + \gamma (h^\mathrm{neq}_i)),
\end{equation}
where the superscript `neq' indicates the non-equilibrium parts $s^\mathrm{neq}_i = s_i - s^\mathrm{eq}_i$ and  $h^\mathrm{neq}_i = h_i - h^\mathrm{eq}_i$. The relaxation parameter $\beta$ is linked to the kinematic viscosity $\nu$ of the fluid through
\begin{equation}
    \beta = \frac{1}{2\left(\frac{\nu}{c_s^2}+0.5\right)}
\end{equation}
with the lattice speed of sound $c_s$, which equals $c_s=\sqrt{1/3}$ in the case of the D3Q27 stencil. Accordingly, the shear part $s_i$ is relaxed similarly as in the BGK model with the relaxation parameter $\beta$ to recover the Navier-Stokes equation in the weakly compressible regime.  By contrast, the non-physical higher-order moments $h_i$ are relaxed dynamically with the relaxation parameter $\gamma$, which is found afresh at each node and time step by minimizing the entropy as described in the work by \citet{Karlin2014}.

\paragraph{Boundary Conditions}

\begin{figure}
    \centering
    \includegraphics{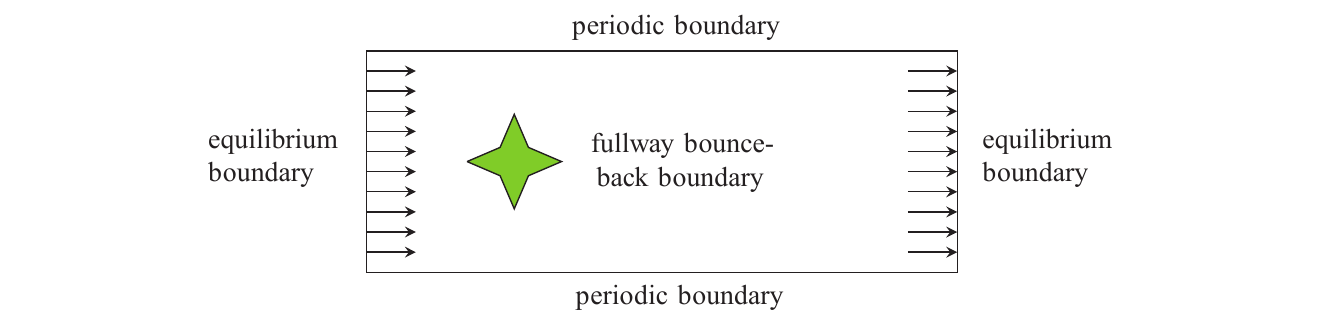}
    \caption{Example of the simulation domain illustrates the used boundary conditions and their application. In 3D, this domain is extruded up from the shown plane, adding an additional set of periodic boundary conditions.}
    \label{fig:boundary_locations}
\end{figure}

For the simulations in this manuscript, stability was the foremost concern to enable hundreds of unsupervised simulations for the multi-solution optimization algorithm. 
For this reason, the periodic boundary, the full way bounce-back boundary, and the equilibrium boundary have been used as shown in figure~\ref{fig:boundary_locations}. 
Preliminary tests, which we conducted prior to the optimization, showed that halfway bounce-back schemes, as well as inlets and outlets relying on neighboring variables such as non-equilibrium boundary conditions, are prone to instabilities at high Reynolds numbers and low resolutions, whereas equilibrium boundary conditions and the fullway bounce-back scheme are extraordinarily stable.

The \emph{periodic boundary} makes populations that are leaving the domain through the boundary on one side reenter at the opposite side. 
The \emph{equilibrium boundary} enforces the equilibrium distribution 
    \begin{equation}
    f_{i}^{\text{eq}}\left(\mathbf{x},t\right) = w_{i}\rho\left( 1 + \frac{\mathbf{u \cdot c}_{i}}{c_{s}^{2}} + \frac{\left(\mathbf{u \cdot c}_{i}\right)^{2}}{2c_{s}^{4}} - \frac{\mathbf{u}\cdot\mathbf{u}}{2c_{s}^{2}} \right)
\end{equation}
on the boundary nodes. 
The equilibrium can be calculated from the macroscopic properties on the boundary and the known speed of sound $c_s$. 
These properties can be chosen to prescribe the behavior of the fluid on the boundary, in particular for the inlet and outlet of the simulation.
The \emph{fullway bounce-back boundary} models the building shape as a no-slip boundary condition. It works by inverting the velocities $c_i$ to $c_{\bar{i}}$ for populations that get streamed to nodes that lie past the boundary. 
These populations will then be streamed back to the nodes they came from in the next streaming step, effectively "bouncing" the populations back
\begin{equation}
    f_{\bar{i}} (\textbf{x}_b,t+2\Delta t) = f_i(\textbf{x}_b,t).
\end{equation}

\paragraph{LBM setup and axial resolution}

\begin{figure}[ht]
    \centering
    \includegraphics{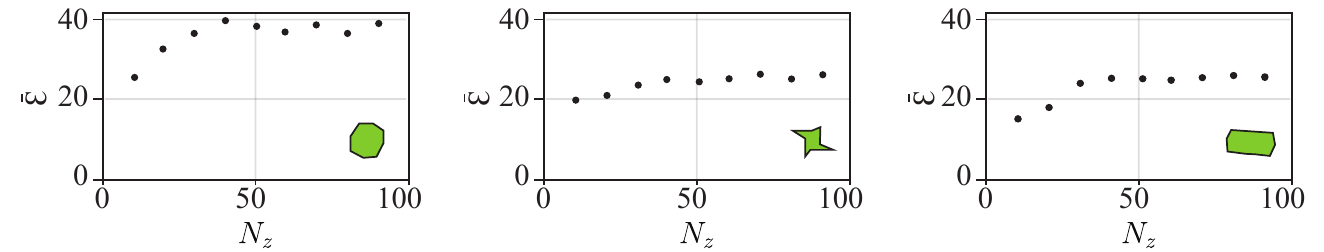}
    \caption{Time-averaged enstrophy over the number of axial grid points $N_z$ for three different shapes.}
    \label{fig:z_extent}
\end{figure}

Here we explain the parameters of the \gls{LBM} simulation. 
The resolution of the two-dimensional domain was $N_x\times N_y = 600\times 300$ and all simulations ran until $t_\mathrm{max}=25$ at inlet Mach number $\mathrm{Ma}=0.075$ and Reynolds number $\mathrm{Re}=3900$. 
In three dimensions, the simulation costs scale linearly with the axial domain size $N_z$. 
For that reason, it is obvious to optimize the number of grid points in axial direction $N_z$ to reduce costs and enable more simulations in the same period of time. 
However, if the axial resolution is too small, the flow will only show two-dimensional behavior. 
Thus, to determine a sufficient resolution in axial direction $z$, an enstrophy study over the number of grid points $N_z$ was performed.
The mean enstrophy $\bar{\mathcal{E}}$ was averaged after a sufficient initialization phase $t>10$.

Figure~\ref{fig:z_extent} depicts the results of three different shapes, revealing that---principally---a resolution of $N_z\approx50$ is sufficient as we observed that larger values do not significantly improve the results with respect to the mean enstrophy $\bar{\mathcal{E}}$.

\subsection{Correlation between 2D and 3D features}
\label{sec:eval:correlation}

\begin{figure}[ht]
    \centering
    \includegraphics{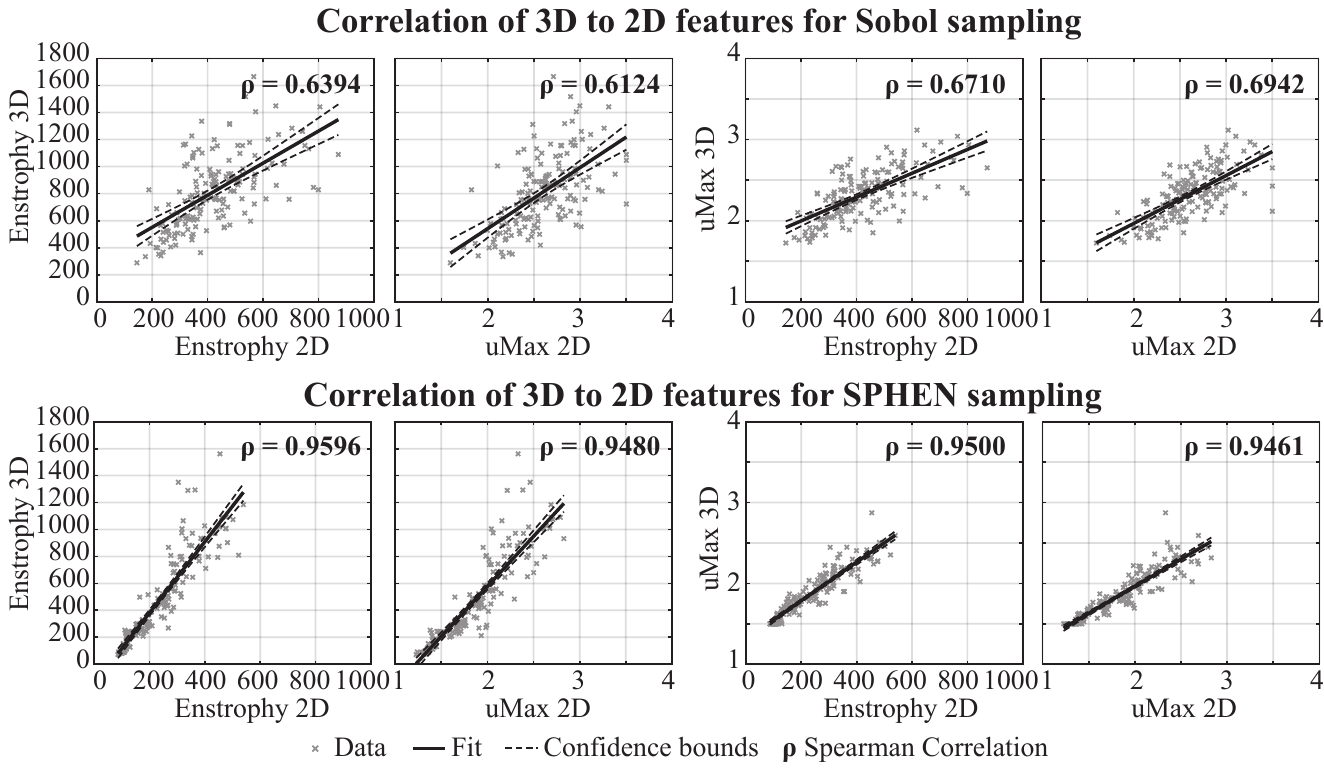}
    \caption{\textbf{Linear regression} of 3D flow features w. r. t. (combinations of) 2D flow features. Top: when sampling with Sobol, bottom: when sampling with SPHEN. Spearman correlation ($\mathbf{\rho}$) is shown for all samples of all 10 replicated experiments.}
    \label{fig:correlation}
\end{figure}

To determine the correlation between 2D and 3D flow features, we compare the 2D and 3D values of $\bar{\mathcal{E}}$ and $\mathbf{u}_{max}$ in the field for sets sampled from a Sobol sequence on the design parameter space to those of an optimized set of shapes. 
The optimization takes place in 2D using Lettuce and \gls{SPHEN}, as explained in the previous sections. 
The expectation is that we can more easily predict 3D flow features for the set of 2D shapes that are optimized towards minimizing $\mathbf{u}_{max}$. 
The high correlation between 2D and 3D features (see figure~\ref{fig:correlation}) shows that the features of the SPHEN sampling set will indeed be easier to predict than those of the Sobol set.

\subsection{Best 2D predictors of 3D features}
\label{sec:eval:bestpredictors}

\begin{figure}[ht]
    \centering
    \includegraphics{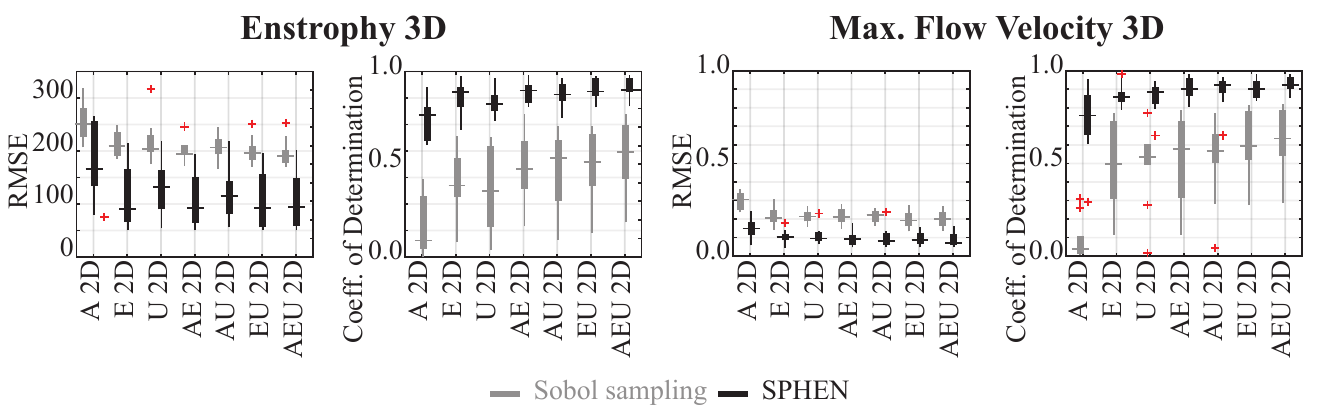}
    \caption{\textbf{Correlation.} Linear model fitted on Sobol versus SPHEN sampling (with 16 chosen representatives). \textit{10 replicated experiments.} Features are abbreviated as follows: floor space (denoted A), enstrophy (denoted E), max. flow velocity (denoted U). RMSE denotes the root mean squared error, coefficient of determination, and outliers in red.}
    \label{fig:linearmodel}
\end{figure}

Here we evaluate the accuracy of the surrogate models that predict 3D features from their 2D counterparts. 
The use of combinations of 2D features might possibly improve model accuracy. 
Before evaluating model accuracy, a linear model is fitted to show how well 3D features are explained by 2D features/combinations. 
Features are combined by using them as multivariate inputs to the linear model.
Results are depicted in figure~\ref{fig:linearmodel}. 

The coefficient of determination between 2D and 3D features for the SPHEN sample set is higher ($\geq 0.9$) than for a Sobol sample set ($\approx 0.6$).
The root mean square error (RMSE) is lower for SPHEN.
This clearly shows that we might expect 3D features to be predictable in the context of evolutionary optimization, at least when using a multi-solution algorithm like SPHEN.
A combination of 2D features (AU, EU, and AEU) produces the highest coefficient of determination, allowing the linear model to explain most of the variability in the 3D response variable based on 2D features.
This shows that making use of a \gls{QD} pre-optimization can help discover features that are better predictors for 3D flow features. 

In the last experiment, it was found that the correlation between 2D and 3D features can be higher for sample sets produced by \gls{QD} and that the combinations of 2D features are potentially the best predictors for 3D features. 
We now use 10-fold cross-validation to determine the predictive accuracy of \gls{GP} models. 
The \gls{GP} models are used to map flow features of 2D shapes to flow features of their extruded 3D counterparts. 
Features are again combined by using them as multivariate inputs to the \gls{GP} model.

\begin{figure}[ht]
    \centering
    \includegraphics{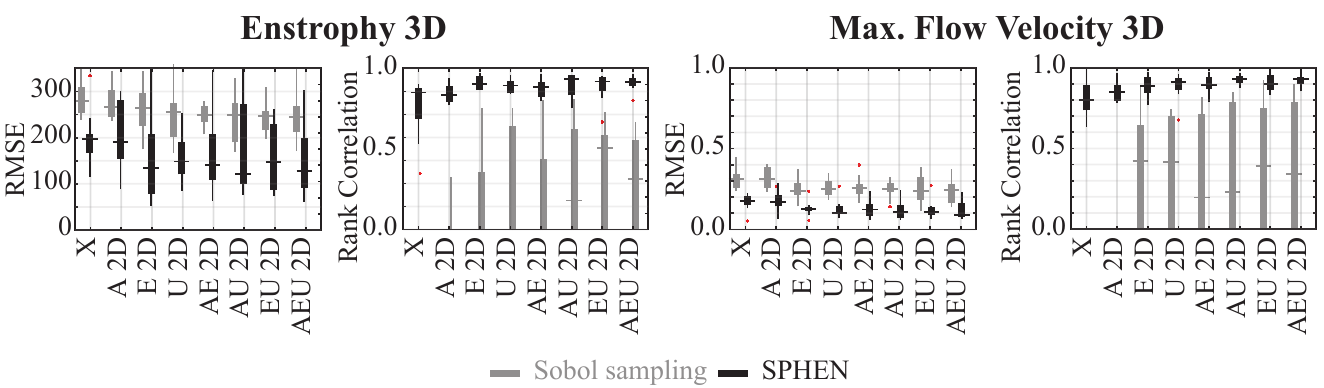}
    \caption{\textbf{Prediction.} \gls{GP} trained on hold-out set (10-fold cross validation) using Sobol sampling versus SPHEN sampling. \textit{10 replicated experiments.} Features are abbreviated as follows: floor space (denoted A), enstrophy (denoted E), max. flow velocity (denoted U). RMSE denotes the root mean squared error, rank correlation by Spearman, and outliers in red.}
    \label{fig:gpcrossval}
\end{figure}

As shown in figure~\ref{fig:gpcrossval}, the rank correlation (Spearman's $\rho$) for \gls{SPHEN} samples is much higher than that of samples generated from a Sobol sequence. 
Especially models trained using 2D floor space, maximum 2D flow velocity, and to a degree 2D enstrophy, seem to create the most accurate predictive models.
We conclude that the diverse, high-performing sampling set produced by \gls{SPHEN} helps us to sample expensive 3D flows more efficiently. 
Prediction based on parameters alone ($\mathbf{X}$ in figure~\ref{fig:gpcrossval}) is possible as well, although not as successful as predictions based on flow features. 
The sample locations can provide accurate surrogate models, allowing us to map 2D shapes to 3D and understanding 3D flow features to some degree from 2D simulations.

A very accurate ranking comparison can be made between the 3D samples of a set produced by SPHEN using a 2D simulation.
This is evidence that we can use a prior 2D optimization step to find better surrogate models to predict 3D features.
We can use these 2D shape sets and the accompanying models to find a suitable initial solution set for 3D optimization.
The surrogate models can also potentially be used to map further 2D simulations to 3D, allowing a quick scan of 3D's predicted flow features before deciding to run a simulation.

\subsection{Example}
\label{sec:eval:example}

\begin{figure}[ht]
    \centering
    \includegraphics{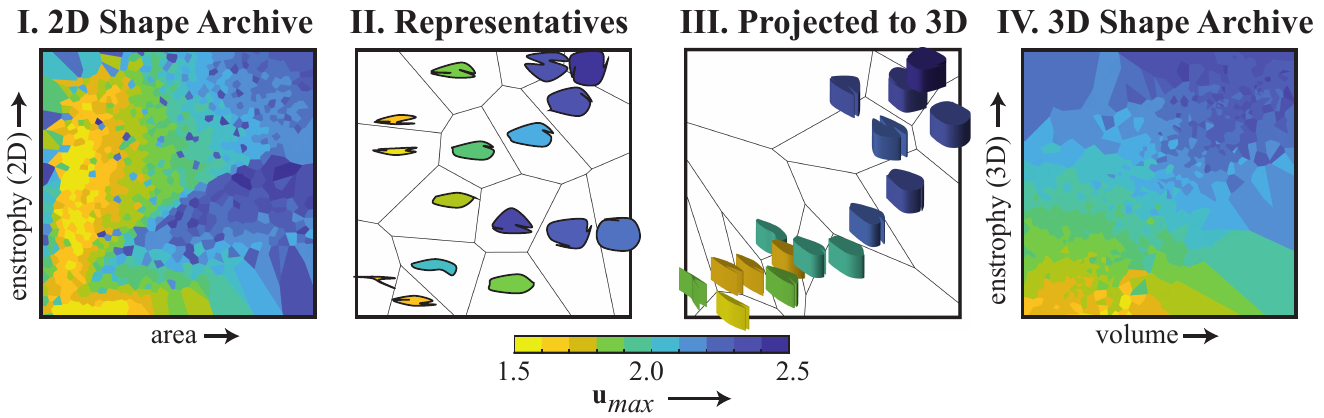}
    \caption{\gls{SADIM} uses \gls{SPHEN} to create 1024 2D shapes (I). Representatives of these shapes are selected (II) and projected to 3D using extrusion and simulation (III). The representatives are then used to predict the feature locations of all remaining 1008 3D extruded shapes (IV).}
    \label{fig:mapped}
\end{figure}

Finally, we show an example of a \gls{SADIM} run in more detail. 
Figure~\ref{fig:mapped} shows the SPHEN archive that contains 1024 building footprints (I). 
The archive is reduced by rebuilding the archive with a maximum number of 16 representative 2D building footprints (II).
This selection method produces a high-performing set of footprints, representative of the floor space-enstrophy variations.
After extruding the shapes to 3D, their flow features are re-determined in simulations with Lettuce (III).
Surrogate models are trained on the flow around these extrusions and used to predict the projection of all 1024 2D building footprints to 3D building designs (IV).
If an architect would be interested in solutions that have a high volume/floor space but low induced wind speeds, the diversity of solutions in orange is at their disposal. 
We can easily see that the information in the figure can provide useful information to architects and engineers at an early design stage.

\subsection{Discussion}

We showed that \gls{QD} provides more appropriate samples than using a pseudo-random space-filling sampling method like Sobol.
3D flow features are more correlated to their 2D counterparts, which was shown in section~\ref{sec:eval:correlation}.
In Section~\ref{sec:eval:bestpredictors}, we also found that we can more easily predict 3D features based on their 2D counterparts using linear or \gls{GP} models.
The rank correlation of 3D features to single 2D features or combinations of features was much higher for those samples that were created with \gls{QD}.
We then showed in section~\ref{sec:eval:example} that we can project large \gls{QD} archives from 2D to 3D solutions using the \gls{GP} surrogate models.
Therefore we conclude that we can indeed produce better \gls{ML} models by using \gls{QD} training data instead of using common sampling techniques like Sobol sequences and that we can map high-performing 2D to 3D shapes using the method we introduced.

The work in this article serves as a proof of concept in which we allow ourselves a number of simplifications, such as leaving out the ground below the 3D building and its roof. 
In addition, the aspect ratio of the building certainly has an influence on the wake of the flow. More detailed simulations of single buildings would also consider turbulent inflows or boundary layer modeling.
As this work's focus is on predicting 3D flow features from 2D counterparts, we do not provide a full validation of the produced 3D extrusions.
We rather focus our validation on the accuracy of predicted 3D flow features.
Despite these simplifications, we believe that, although there is no variation in the 3D shapes w.r.t. the $z$ dimension, this study describes how a multi-solution optimization can provide insights before further improving the results by simulations of selected entire building designs with more detailed resolutions. 

Due to the qualitative differences in turbulent flow when comparing 3D to 2D flow, we have given evidence that the approach might work even for very complex domains because we can predict the (ranked) differences of solutions in 3D flow based on their differences in 2D flow.

\section{Conclusions}
\label{sec:conclusions}

To tackle wind nuisance around buildings early on in the design process, we introduced an efficient method to retain a large variety of 3D building designs, obtained via optimization w. r. t. reducing the maximum wind load around the building.
We showed that a surrogate-assisted approach to quality diversity (\gls{QD}) can be used to find good sampling locations of an expensive 3D flow simulation in order to train surrogate models that can map 2D to 3D flow features for a diverse building set.
In our evaluation, we showed that we can produce better \gls{ML} models, \gls{GP} specifically, by using training data generated with \gls{QD} algorithms, rather than using space-filling sampling methods.
Especially in the context of generative design and optimization, it makes sense to use a divergent optimization technique to generate training data.
We showed that we can map the features of flow around high-performing 2D shapes to those of flow around their extruded 3D counterparts.
Using \gls{SADIM}, surrogate-assisted dimensionality mapping, we can map the \gls{QD} archive holding the 2D shapes to one that holds 3D shapes.

Our assumption is that the archive with 3D shapes is a good starting point for further 3D optimization.
In future work, we will extend the method with a full 3D search, based on those initial starting locations produced by \gls{SADIM}. This extension will necessitate adding more parameters to the 3D shape, which can deform the building in the z-direction.
The accuracy of the surrogate models will need to be evaluated for unknown samples in a 3D search.
\\ \\
The archives produced by \gls{QD} can be used as an initial population for a global or local search in the 3D flow regime, potentially improving the efficiency of the search.
They can also be used as a source of informed inspiration for engineers. 
Specifically, the use case showed this for high-rise building designs that cause low wind nuisance to the environment while maximizing the floor space. 
Providing an intuition to answers like this can support making informed decisions about building design better and earlier. 
It allows engineers to efficiently sweep the design space, select their preferred solutions and then run high-resolution simulations on those to better understand their preferences. 
Finally, this work showed that evolutionary optimization, \gls{QD} specifically, can serve as a data set generator for \gls{ML}.
The ability to generate data sets from scratch enables us to create specific data sets to train or refine efficient \gls{ML} models.

\section*{Acknowledgements}
The authors thank Rudolf Berrendorf and Javed Razzaq for their continuous development and support of the BRSU’s computing cluster.

\section*{Funding}
The computer hardware was supported by the Federal Ministry for Education and Research and by the Ministry for Innovation, Science, Research, and Technology of the state of Northrhine-Westfalia (research grant 13FH156IN6).

\bibliographystyle{apalike}
\bibliography{main} 

\end{document}